\pdfoutput=1

\documentclass[11pt]{article}

\usepackage[final]{acl}

\usepackage{times}
\usepackage{latexsym}
\usepackage{enumitem}
\usepackage{listings}
\usepackage{amsmath}
\usepackage[T1]{fontenc}

\usepackage[utf8]{inputenc}

\usepackage{microtype}

\usepackage{inconsolata}
\usepackage{graphicx}
\usepackage{caption}

\title{Automatic Bug Detection in LLM-Powered Text-Based Games Using LLMs}

\author{
 \textbf{Claire Jin\textsuperscript{1}\thanks{Work done as interns at Microsoft Research}},
 \textbf{Sudha Rao\textsuperscript{2}},
 \textbf{Xiangyu Peng\textsuperscript{3*}},
 \textbf{Portia Botchway\textsuperscript{2}},
\\
 \textbf{Jessica Quaye\textsuperscript{4*}},
 \textbf{Chris Brockett\textsuperscript{2}},
 \textbf{Bill Dolan\textsuperscript{2}}
\\
\\
 \textsuperscript{1}Carnegie Mellon University,
 \textsuperscript{2}Microsoft Research,
 \textsuperscript{3}Salesforce Research,
 \textsuperscript{4}Harvard University
 \\
 \\
\href{mailto:claireji@andrew.cmu.edu}{claireji@andrew.cmu.edu}
 }

\begin{document}
\maketitle
\begin{abstract}

Advancements in large language models (LLMs) are revolutionizing interactive game design, enabling dynamic plotlines and interactions between players and non-player characters (NPCs). However, LLMs may exhibit flaws such as hallucinations, forgetfulness, or misinterpretations of prompts, causing logical inconsistencies and unexpected deviations from intended designs. Automated techniques for detecting such game bugs are still lacking. To address this, we propose a systematic LLM-based method\footnote{Code and prompts released here: \url{https://github.com/microsoft/llm-game-bug-detection}} for automatically identifying such bugs from player game logs, eliminating the need for collecting additional data such as post-play surveys. Applied to a text-based game DejaBoom!, our approach effectively identifies bugs inherent in LLM-powered interactive games, surpassing unstructured LLM-powered bug-catching methods and filling the gap in automated detection of logical and design flaws.

\end{abstract}

\section{Introduction}
\label{intro}
Text-based computer games are complex, interactive simulations where the game state (location, scenes, and dialogue) is described through text, and players take actions and interact with non-player characters (NPCs) within the game world to achieve an ultimate goal \cite{cote2018textworld}.
NPCs are traditionally pre-scripted, limiting user inputs to text commands \cite{mehta2021exploring} and resulting in stiff interactions. 
More recently, advancements in LLMs have enabled a paradigm shift in text-based games, allowing for dynamic interactions between players and NPCs \footnote{https://www.aidungeon.com/},
with major companies\footnote{https://developer.nvidia.com/ace} releasing LLM-powered ``character engines''\footnote{https://inworld.ai/character-brain} for adaptive, unscripted NPCs with backstories, goals, emotions, and free-form conversations. 
These capabilities extend beyond NPCs to include other game design aspects such as level generation \cite{kumaran2023end, taveekitworachai2023ChatGPT4PCG, tsai2023large}.

The plotline-flexibility, NPC adaptability, and text input freedom of LLM-powered games, however, also creates opportunities for logical inconsistencies, hallucinations, and memory-loss mistakes. To ensure the playability of LLM-powered games, automatic detection of these bugs is essential. Particularly, in our work, we focus on logical and game balance bugs. Logical bugs,  
per \citet{zheng2019wuji}, involve flaws in the logic implementation, resulting in unexpected outcomes without interface crashes, while game balance bugs are deviations from the designers’ intentions, potentially making games overly challenging or easy \cite{zheng2019wuji}. 
Such bugs are rare in traditional games due to the predefined nature of both NPC dialogues and plotlines. Hence, most previous work on automated bug detection has focused on graphics bugs, crashes, or freezes \cite{azizi2023automatic,macklon2023automatically,varvaressos2017automated,alnassar2023questville}, with little attention to logical and game balance bugs. Moreover, traditional detection of these bugs relies mainly on sentiment analysis of player feedback surveys and game designer discretion \cite{zheng2019wuji}. These surveys require extra effort to gather, but offer limited insights for complicated issues in games \cite{Su2020ComprehensiveRA}.

To address this gap, we develop an LLM-powered method to automatically detect player pain points and associated game logic and game balance bugs from game logs.
Unlike player surveys, 
which hinge on players' post-game recollection,
game logs (example in Fig~\ref{fig:game step}a), are automatically generated during gameplay, offering richer insights by capturing real-time player actions and dialogue within the game’s progression. However, 
identifying 
game logic or balance bugs from logs of text-based games is an intricate task. 
It requires synthesizing multiple sources of information (game logic/goals and player progression), inferring intention
from player action/utterance, and identifying subtle causal relationships between player behavior and game outcome.
Thus, this task remains challenging, even for human experts.
We seek to leverage LLM's language reasoning
to address this challenge. Feeding game logs directly into an LLM, however, does not yield meaningful identification of logic and balance bugs due to the 
complexity of this task. We therefore develop a structured approach that leverages the game designer's intended gameplay to guide an LLM in mapping diverse gameplay attempts to a trackable unified framework. This enables the aggregation of gameplay experiences across players for identifying potential game logic and balance bugs. 

Our two main contributions are:
\begin{itemize}[noitemsep,topsep=0pt]
    \item We present a novel method of automatically assessing LLM-powered text-based games for logical and game balance bugs arising from LLM-driven NPC dialogue and plotlines (Section~\ref{methods}). To the best of our knowledge, this is the first method for this task.
    \item Our method provides objective, quantitative and scalable assessments on the difficulties of each game part in a text-based game (Section~\ref{results}).
\end{itemize}

\section{DejaBoom! LLM-powered Game}
\label{dejaboom}
We test our method on the recently published text-based mystery game ``DejaBoom!'' \cite{peng2024player}. This game utilizes a plot-constrained GPT-4 to generate all in-game text, including game state descriptions, outcomes of player actions, and NPC dialogue, is LLM-generated \cite{cote2018textworld}.

The game begins with the player waking up at home in a village, experiencing ``deja vu'' about an explosion. To solve the game, the player must obtain a bomb disposal kit and locate the bomb before diffusing it. The player
must explore the village, interact with NPCs and items to understand how to solve the mystery. Some NPCs and items can be useful for solving the mystery, but others are red herrings.

The bomb detonates after a fixed number of gameplay steps (N=30), where each step is a single in-game player action/utterance, causing the player to relive the day until diffusing the bomb. They retain memory across explosions, but the game world and NPCs are reset after each explosion.\footnote{Further game details are in Appendix \ref{appendix:dejaboom}.}

\citet{peng2024player} released game logs for 28 players, including all player text inputs and game engine outputs during their gaming sessions. These logs include details of player actions and utterances, NPC utterances, and current game state information such as location, inventory, and outcomes of player actions.\footnote{See Appendix \ref{appendix:log} for a sample log.} Additionally, they published post-study survey responses from the players, offering additional insights into their experiences and feedback on the game.

\section{Methods}
\label{methods}
\begin{figure}[ht]
    \centering
    \includegraphics[width=0.5\textwidth]{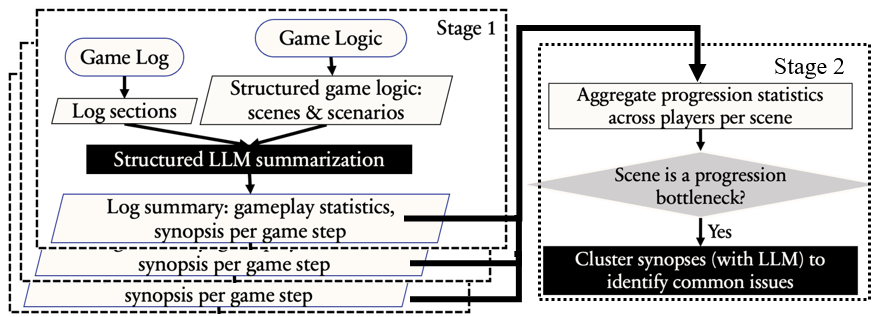}
    \caption{Our automated bug detection procedure. 
    }
    \label{fig:flowchart}
\end{figure}

\begin{figure*}[ht]
    \centering
    \includegraphics[width=\textwidth]{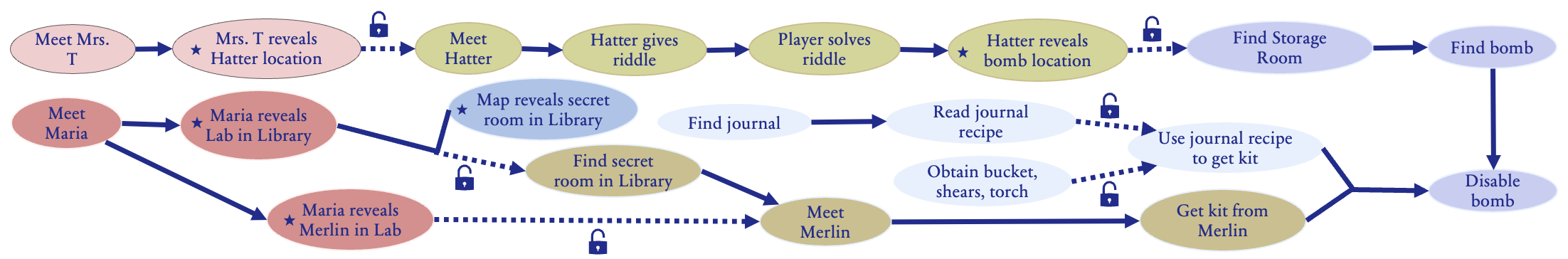}
    \caption{DejaBoom logic graph. Nodes represent scenes and colored groups indicate scenarios. Arrows indicate order of completion. Merging arrows: only one of the tail nodes is required to proceed to the head node. Dotted arrows: a new location, NPC, or item should be unlocked.}
    \label{fig:game graph}
\end{figure*}

\begin{figure}[ht]
    \centering
    \includegraphics[width=0.5\textwidth]{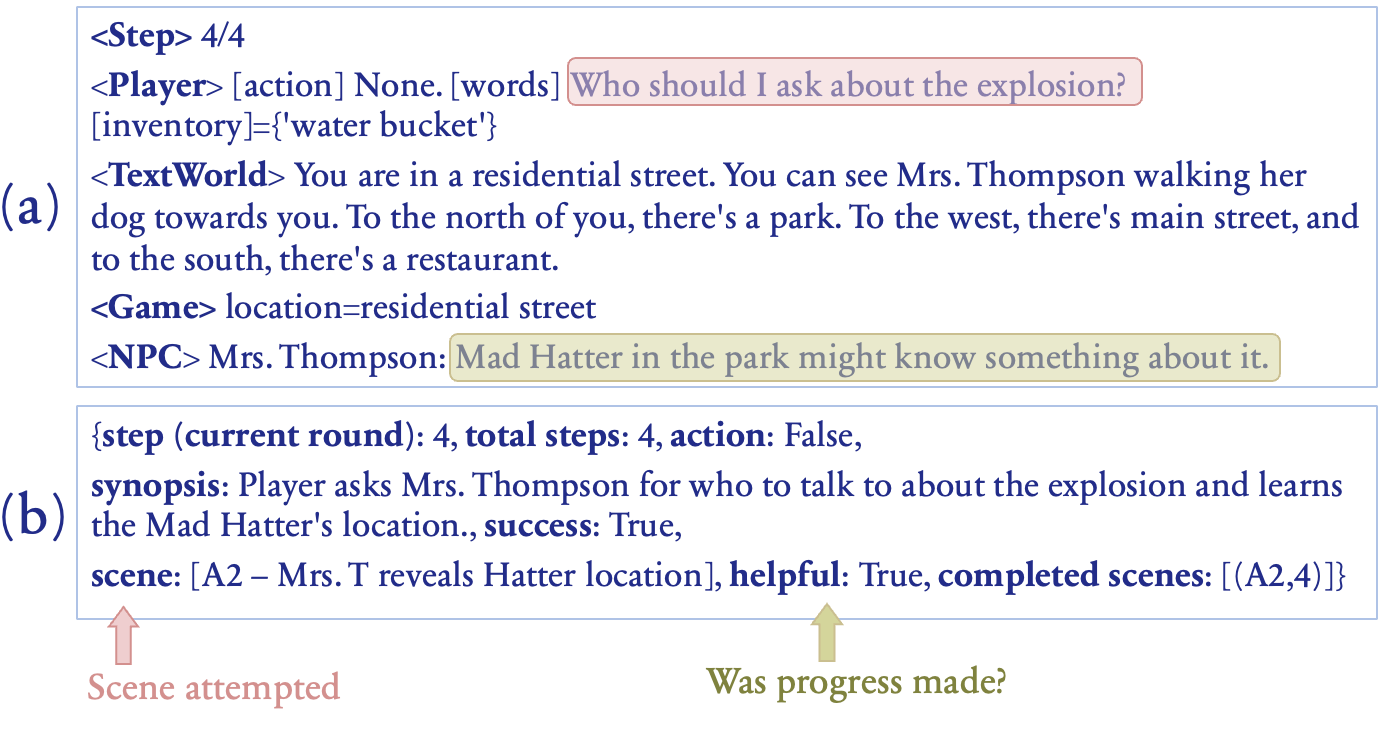}
    \caption{(a) A gameplay step from a section inputted to the LLM. (b) LLM-generated summary for (a).}
    \label{fig:game step}
\end{figure}

Our automated bug detection procedure takes players' game logs and the designer's intended game logic as input, identifying bugs using an LLM-powered two-stage procedure (Fig~\ref{fig:flowchart}). 
Stage 1 uses a structured approach to prompt the LLM to map various player attempts recorded in a game log to the designer's intended progression roadmap and extract information on the player's progression and gameplay experience into a standardized summary that is comparable across players and gameplay steps (Section~\ref{methods:summarization}). Stage 2 uses an LLM to aggregate these summaries across players to quantify the difficulties of each game progression unit, identify common progression bottlenecks, and pinpoint possibly associated logic and game balance bugs (Section~\ref{methods:statistics}). 
We use GPT-4 \cite{openai2023gpt4}, the current best of breed model, for all our LLM needs.

\subsection{
Stage 1: Structured alignment and summarization of gameplay progression}\label{methods:summarization}

To help the LLM establish a progression roadmap that reflects the intended game flow, we segment the intended game logic into ``scenarios’’. 
Each scenario encapsulates a plotline segment with a specific aim necessary for progression and is subdivided into ``scenes’’. Each scene has a distinct goal, with an immediate success indicator.
Each scene goal requires at least one action or utterance from the player to complete. The game logic graph of DejaBoom! (Fig~\ref{fig:game graph}) illustrates this structure. 
The node relationships in the graph are provided to the LLM for aligning player progress recorded in a game log with the intended game flow. This step gives the designer's intended game flow a clear structure, allowing us to better leverage the LLM's reasoning capabilities.

To prepare game logs for alignment, we first preprocess the raw logs by removing extraneous details and redundant information, and adding tags for NPC utterances, player input, and game world feedback.\footnote{Examples in Appendix \ref{appendix:log}).} The preprocessed logs are then segmented into sections of two consecutive gameplay steps, each comprising a single player action/utterance (example in Fig~\ref{fig:game step}a). Presenting two steps per section provides the LLM with additional temporal context compared to single-step sections, enhancing its ability to understand sequential progression. 
Due to GPT-4's token limits, we were unable to use three-step sections.

To map the progression in a game log to the intended game flow, the segmented game sections from each play session, ordered by gameplay steps, are provided to the LLM sequentially, along with the game logic, as input. The LLM is instructed to generate a structured summary (Fig~\ref{fig:game step}b) for each gameplay step, outlining the player’s action and outcome in a 1-sentence synopsis, mapping the player's progress to a scene of the game logic graph, recording completed scenes, and assessing whether the step contributes to progression. The summaries across players are used to compute difficulty metrics for each scene in Stage 2.

To train the LLM for this task, we employ a few-shot learning protocol, with 60 curated summaries as in-context training examples, covering a wide range of player experiences. \footnote{Examples are included in Appendix \ref{appendix:examples}.} All LLM-generated summaries underwent an automated legality check to ensure temporal consistency within each log \footnote{Details can be found in Appendix~\ref{appendix:consistency}}.

\subsection{Stage 2: Identify pain points and causes}\label{methods:statistics}

To assess difficulty at a scene, we aggregate the summaries produced from Stage 1 (Fig \ref{fig:game step}b) across players for each scene, and compute scene completion rates.
Scenes with low completion rates or significant drops in completion rate from the preceding scene are flagged as potential pain points. Scenes with unusually high completion rates may also be flagged, as this may also be a deviation from the designer's intent.

For each flagged scene, we cluster the 1-sentence synopses generated in Stage 1, which summarize player actions, experiences, and outcomes, across players using an LLM.
By grouping synopses with similar semantic content, the LLM forms clusters representing various types of common experiences encountered in gameplay during a particular scene. These clusters reveal common player attempts, outcomes, and obstacles, facilitating the identification of specific causes of hindrances.

\section{Results}
\label{results}
\begin{figure}[ht]
    \centering
    \includegraphics[width=0.49\textwidth]{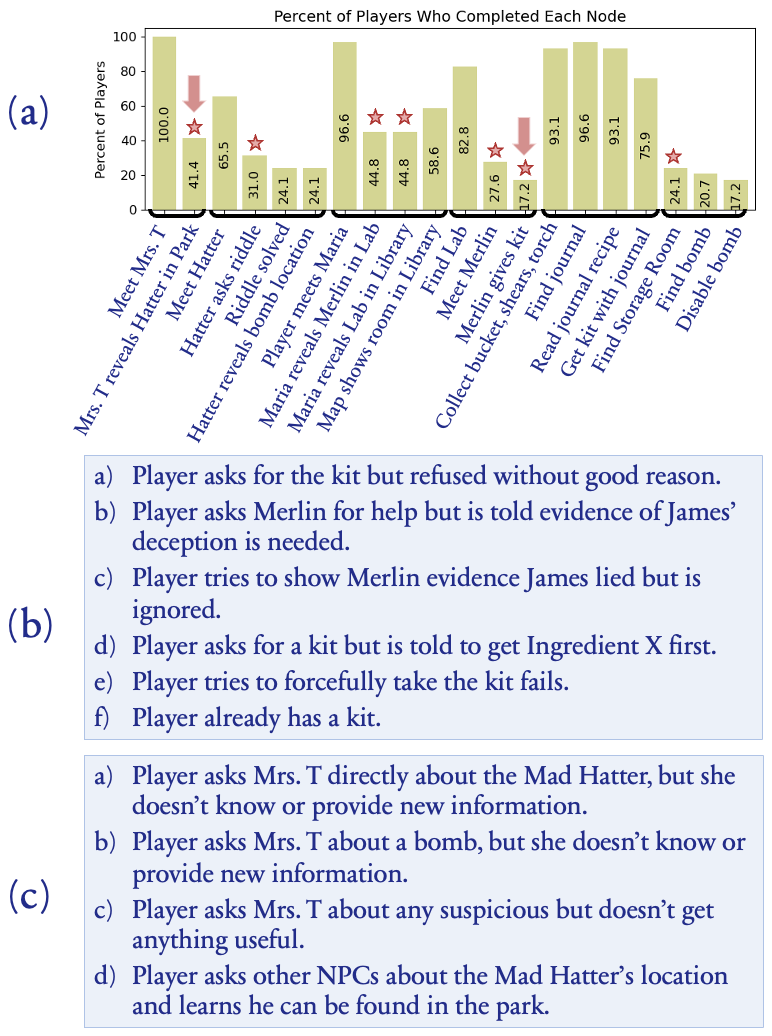}
    \caption{(a) Completion rate per scene. Black brackets group scenes forming a scenario. Stars mark potential pain point scenes. (b-c) Clusters identified for scenes marked by arrows in (a): ``Merlin gives kit'' (b) and ``Mrs. T reveals Hatter in Park'' (c).}
    \label{fig:results}
\end{figure}
\subsection{Identifying progression bottlenecks}
Scenes with a low completion rate ($<20\%$) or noteworthy completion rate decline ($<1/2$ of the preceding scene) within their corresponding scenarios are flagged as potential pain points. Out of 20 scenes, 7 are flagged (Fig \ref{fig:results}a). For example, the ``Merlin gives kit'' scene, despite not being the ultimate goal of the game, has the lowest completion rate (17.2\%) among scenes, matching that of the final scene, ``Disable bomb’’. 
The ``Mrs. T reveals Hatter in Park’’ scene has a completion rate less than half of the preceding scene, indicating a bottleneck in progression.

\subsection{Identifying common causes of bottlenecks} 

We present the resultant clusters from Stage 2 for two of the flagged scenes below and present the rest in Appendix \ref{appendix:clusters}.

For the "Merlin gives kit" scene (Fig ~\ref{fig:results}b), clusters (a)-(d) expose logic bugs where NPC Merlin's behavior differed from expectations. He mistakened NPC James and Ingredient X, both intended as red herrings, as necessary for obtaining the bomb disposal kit, hindering player progress. Clusters (e)-(f) reveal player confusions during game exploration, particularly when players try to get a kit from Merlin despite already having one. This suggest a potential balance bug if disproportionately many players waste time trying to acquire an unnecessary item due to the game design.

For ``Mrs. T reveals Hatter in Park’’ (Fig ~\ref{fig:results}c), clusters (a)-(c) reveal a balance bug, as Mrs. Thompson only accepts the keyword ``explosion'' as the unlocking criterion in player queries, needlessly increasing the scene difficulty by excluding related terms like "bomb".
Cluster (d) uncovers a logic bug in the LLM game engine, which mistakenly allows NPCs other than Mrs. T to disclose the Hatter's location to the player, yet the player cannot unlock the Hatter because the unlocking condition is Mrs. T's disclosure.

To verify the bugs detected by our method against those identified via survey and manual parsing, we include a survey-based ground truth along with an inspection of game logs. Players reported 7 bugs in the survey, 5 of which were detected by applying Stage 2 of our method to the flagged scenes (Fig~\ref{fig:results}a). However, upon applying the clustering step in Stage 2 of our method to additional scenes, the remaining two bugs were also revealed. Of the 7 potential pain point scenes identified by our method, 5 were corroborated by the survey. The remaining two scenes concerned NPC Maria, who received general complaints in the survey without specific bug reports. These findings indicate that our method aligns with insights from player feedback surveys but offers greater specificity. Consequently, our method not only eliminates the need for manual feedback collection and parsing, but also provides deeper insights into player experience beyond typical survey feedback. Further details of the survey and comparison analysis can be found in Appendix \ref{appendix:ground-truth-comparison}.

\subsection{Ablation Studies}
\label{results:ablation}
\begin{table*}[!htpb]
\centering
\begin{tabular}{cccc}
\hline
\textbf{Method} & \textbf{\# of game parts} & \textbf{Easy aggregation} & \textbf{Reveals bug causes?}\\
\hline
Naive & 661 & No & No\\
No logic & 400 & No & Potentially\\
Ours & 20 & Yes & Yes\\\hline
\end{tabular}
\caption{For each method (ours and two ablations), we indicate the number of unique game parts the method found prior to any semantic post-processing. We also indicate whether the game parts identified are easy to aggregate across gameplay sessions, and if the method reveals specific causes of bugs via tracing game parts back to specific player experiences.}
\label{tab:ablation}
\end{table*}

We conduct two ablation studies for comparison: (1) a naive method lacking both our game logic (scenarios/scenes) and summarization structures, and (2) a method lacking the logic structure but retaining the summarization structure. In all methods, we seek to extract parts of the game (i.e. “game parts”, for example, learning the bomb’s location) in which there are logic or balance bugs, then to identify the bugs. In (1), the LLM is asked to identify ``game parts'' that the player struggled with and to write a prose summary for each game section. In (2), the LLM follows the summary structure in Fig \ref{fig:game step}b but determines game parts and assigns gameplay steps to ``game parts'' on the fly instead of to our pre-defined scenes, which are consistent across gameplay sessions. The same log sections are provided in all experiments, with new in-context examples per experiment. Prompts and examples for each experiment are in Appendix \ref{appendix:ablation}.

Both methods show significant shortcomings. Particularly, quantifying the difficulty of ``game parts'' is challenging due to the lack of a standardized set of  progression units for aggregating player experiences. 
Without scenarios and scenes, which function as landmarks in the game, to map player progression to, the LLM often reports the player's game step (i.e. a single action or utterance from the player and the immediate consequence) as the "game part". The game step, however, is specific to a particular gameplay session, thus cannot be compared across players or even across sessions for the same player. 
As shown in the column ``\# of game parts'' in Table \ref{tab:ablation}, the two ablation methods report a much larger number of syntactically unique game parts that players progressed through than our method. Upon inspection many of these “unique” parts are overlapping with respect to the game sequence.
For instance, ``acquiring the disposal kit from Merlin'' and ``acquiring the disposal kit through non-violent means'' are synonymous, but this is not obvious without game context.
Extensive, likely manual, post-processing with game knowledge would be required to reconcile such game parts to an appropriate level of granularity for accurate quantification of difficulty.
Additionally, many identified game parts are unrelated to game logic or overly vague. For instance,``attempting to sit on a non-existent bench'' is a user error irrelevant to game logic, while 
``lack of progress towards objectives'' is overly generic. Again, manual post processing with game context would be required to discard such game parts.
These challenges therefore make it difficult to determine if a struggle for one player is recurrent across players, highlighting the importance of alignment with a consistent game logic structure for tracking and aggregating player progression.

Ablation method (1) also presents extra challenges in extracting common causes of player struggles compared to (2). The prose summary of (1) requires laborious parsing and reasoning to extract relevant information for determining causes of a struggle point, which can sometimes be infeasible.
This demonstrates the effectiveness of our structured summarization in Stage 1 of our method. 
Further discussion on shortcomings of (1) and (2) are in Appendix \ref{appendix:ablation-results}.

\section{Conclusion}
\label{conclusion}
Our work introduces a novel pipeline for automatically identifying logic and balance bugs
behind player pain points
from game logs of text-based adventure games. It is the first automated bug detection method for developing high quality, LLM-powered games. By effectively extracting insights from rich content of game logs, it provides an objective and quantitative platform for bug identification, enhancing game design efficiency.

Future directions include testing this work on more complex text-based games and multimodal games. Additionally, the output of our framework may also be fed directly into an LLM-powered game to explore ways the game can automatically adjust without explicit game designer intervention.

\section*{Limitations}
\label{limitations}
The LLM used in this study was Open AI’s GPT-4. This model was chosen as a SOTA model. Other LLMs are readily to be used in our pipeline. The results may have some LLM-dependent variation. Given the lack of other works in this area at the time of conducting this research, we only perform ablation comparisons with potential alternative methods, but not any published methods.

The DejaBoom! game was designed and played using the English language, the interactional behaviors observed were those of English-speaking players who were living and working in the United States, and the game logs were in English. While our framework generalizes to games and game logs in other languages, its effectiveness may depends on the LLM's ability to operate in languages other than English.

While our framework has only been tested on the DejaBoom! game, we anticipate a smooth transfer to other adventure games. 
Pain points in all adventure games typically manifest as progression bottlenecks, and our framework is specifically designed to capture these bottlenecks and their underlying causes. Our framework can be readily applied to other adventure games by making modifications to parse the game logic and logs. However, additional adaptations may be necessary for other types of games.

\section*{Ethics Statement}
\label{ethics}
To power the Dejaboom game, an unfiltered LLM was used due to initial difficulties in getting a filtered model to handle inputs pertaining to explosive devices.
No additional filtering was implemented in our code, our working assumption being that the constraints of the game would themselves function effectively as baseline harm mitigation for the purposes of experimentation.  
Participants were advised that that they might be accidentally exposed to harmful language, but we are unaware of any such incidents. 
In a public-facing game scenario, it will nevertheless be necessary to implement harm mitigations tailored to the purpose of the game.
All participant data was fully anonymized.

\section*{Acknowledgements}
\label{acknowledgements}
We would like to thank the anonymous reviewers for their valuable comments.
We also thank Qunhua Li, Yonatan Bisk, and Daniel Anderson for their critical feedback and discussions.

\bibliography{custom}

\appendix
\section{Appendix}
\label{appendix}
\subsection{Dejaboom! Expanded}
\label{appendix:dejaboom}
\begin{figure*}[ht]
    \centering
    \includegraphics[width=\textwidth]{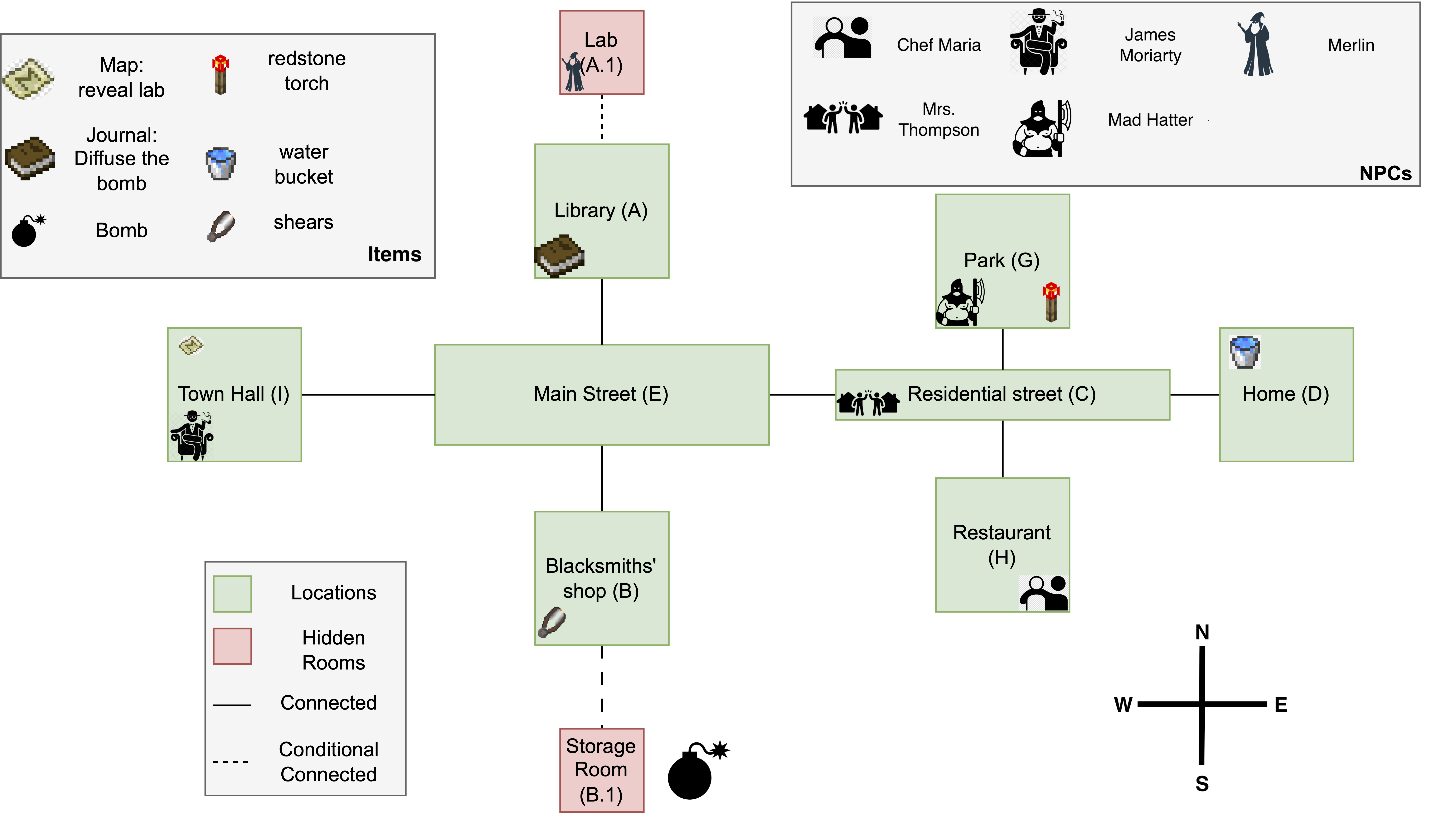}
    \caption{Dejaboom! game layout. A map of the village where the game takes place, showing the locations, objects, and NPCs. The player begins the game from home and their goal is to diffuse the bomb before it explodes again.}
    \label{fig:dejaboom}
\end{figure*}

In this section, we provide abbreviated details of the DejaBoom! game. A full description of the game can be found in \citet{peng2024player}. Fig ~\ref{fig:dejaboom} shows a map of the game world.

The village's has six main locations: the home, park, restaurant, blacksmith's shop, library, and town hall.
There are also two secret rooms: a lab in the library and a storage room in the blacksmith's shop, where the bomb is hidden. These rooms become accessible only after the player meets certain conditions. 

The NPCs and their roles are Merlin, who can provide a bomb disposal kit; Chef Maria, who can unlock Merlin's location; the Mad Hatter, who can unlock the bomb location; Mrs. Thompson (Mrs. T), who can unlock the Mad Hatter's location; James Moriarty, who is a red herring and does not unlock anything. 

The game includes several items the player can interact with and put in their inventory. These are (1) four ingredients located around the village that automatically form a bomb disposal kit when all collected and (2) a map located in the town hall unlocking the hidden rooms. 

To make the disposal kit from ingredients, the player must collect the water bucket from the home, the redstone torch from the park, the shears from the blacksmith's shop, and read the recipe in the journal in the library (not necessarily in this order). Alternatively, as mentioned before, the player can obtain a bomb disposal kit from NPC Merlin.

To beat the game, players need to attain locate the bomb and acquire the bomb disposal kit before diffusing the bomb with the kit.

\subsection{Sample Game Logs}
\label{appendix:log}

In this section, we display a portion of a raw game log and the corresponding cleaned log. We also discuss what each section of the cleaned log means and which information was removed during the cleaning process.

\subsubsection{Raw Game Log}
\begin{lstlisting}[numbers=none]
2023-07-13 16:06:08,640 MainThread INFO LLM:gpt-4-32k
2023-07-13 16:06:08,640 MainThread INFO Azure?:True
2023-07-13 16:06:08,640 MainThread INFO Reset step?:30
2023-07-13 16:06:08,640 MainThread INFO rate_limit_per_minute:1200000
2023-07-13 16:06:08,641 MainThread INFO <Game reset> 0
2023-07-13 16:06:08,641 MainThread INFO <location> home
2023-07-13 16:06:08,659 MainThread INFO <Game> -= Welcome to DejaBOOM! =- 

Game instruction (Please read carefully):
A text game, also known as a text-based game or interactive fiction, is a type of game where the player interacts with the game world through written text. The game provides descriptions of environments, characters, and situations, and the player types commands to perform actions and progress through the story.

In order to engage with the game, you are only allowed to use one of these three types of commands. 
1. Verb-object format. Commonly used verbs are "go", "take," "read", and "open." Example: "pick up watch", "go north" or "go to park"
2. Check inventory: "i"; read instructions: "help".
3. To talk with NPCs, type what you want to say directly. Example: type "how are you?".

Please perform ONLY one action OR speak once at a time. Do not do both of them or multiple actions.

-=GAME START=-

-= Home =-

You wake up in your bedroom. As you look around the room, everything seems familiar, yet somehow different. Suddenly, a sense of deja vu washes over you, and you remember the explosion in the village that occurred yesterday. You realize that you are reliving the same day again. You feel a sense of urgency to stop the explosion from happening again.
You see a wooden table standing in the center, with a water bucket placed on top. The atmosphere is quiet and uncluttered.
The door to the residential street is on your west.
2023-07-13 16:08:54,455 MainThread INFO <Game step> 1/1
2023-07-13 16:08:54,455 MainThread INFO <Player> go west
2023-07-13 16:08:59,813 MainThread INFO <TextWorld> You are in a quiet residential street. You can see Mrs. Thompson walking her dog towards you. To the north of you, there is a park. To the west, there is a main street, and to the south, there is a restaurant.
2023-07-13 16:08:59,817 MainThread INFO <Player_processed>[action] go west. [words] None. [inventory]={}
2023-07-13 16:09:08,657 MainThread INFO <Game> location = residential street    inventory={}    [id]Mrs. Thompson    [words]Hello there! It's a lovely day, isn't it? How are you doing today?        Scene: You are in a quiet residential street. You can see Mrs. Thompson walking her dog towards you. To the north of you, there is a park. To the west, there is a main street, and to the south, there is a restaurant.
2023-07-13 16:10:09,919 MainThread INFO <Game step> 2/2
2023-07-13 16:10:09,919 MainThread INFO <Player> Internally I know she is doomed so I move past her to the town.
2023-07-13 16:10:14,694 MainThread INFO <Player_processed>[action] None. [words] Internally I know she is doomed so I move past her to the town.. \n [inventory]={}
2023-07-13 16:10:23,327 MainThread INFO <Game> location = residential street    inventory={}    [id]Mrs. Thompson    [words]Oh, you seem to be in a hurry. Well, do take care and have a great day!        Scene: You are in a quiet residential street. You can see Mrs. Thompson walking her dog. To the north of you, there is a park. To the west, there is a main street, and to the south, there is a restaurant.
2023-07-13 16:11:05,612 MainThread INFO <Game step> 3/3
2023-07-13 16:11:05,613 MainThread INFO <Player> go west
2023-07-13 16:11:09,677 MainThread INFO <TextWorld> You are in the Main Street. To your north lies a library, to the west is the town hall, and to the south is a Blacksmith's shop.
2023-07-13 16:11:09,680 MainThread INFO <Player_processed>[action] go west. [words] None. [inventory]={}
2023-07-13 16:11:16,761 MainThread INFO <Game> location = main street    inventory={}        Scene: You are in the Main Street. To your north lies a library, to the west is the town hall, and to the south is a Blacksmith's shop.
\end{lstlisting}

\subsubsection{Cleaned Game Log}
\begin{lstlisting} [numbers=none]
-= Home =-

You wake up in your bedroom. As you look around the room, everything seems familiar, yet somehow different. Suddenly, a sense of deja vu washes over you, and you remember the explosion in the village that occurred yesterday. You realize that you are reliving the same day again. You feel a sense of urgency to stop the explosion from happening again.
You see a wooden table standing in the center, with a water bucket placed on top. The atmosphere is quiet and uncluttered.
The door to the residential street is on your west.
<Step> 1/1
<Player> [action] go west. [words] None. [inventory]={}
<TextWorld> You are in a quiet residential street. You can see Mrs. Thompson walking her dog towards you. To the north of you, there is a park. To the west, there is a main street, and to the south, there is a restaurant.
<Game> location=residential street
<NPC> Mrs. Thompson: Hello there! It's a lovely day, isn't it? How are you doing today?
<Step> 2/2
<Player> [action] None. [words] Internally I know she is doomed so I move past her to the town..  [inventory]={}
<TextWorld> You are in a quiet residential street. You can see Mrs. Thompson walking her dog towards you. To the north of you, there is a park. To the west, there is a main street, and to the south, there is a restaurant.
<Game> location=residential street
<NPC> Mrs. Thompson: Oh, you seem to be in a hurry. Well, do take care and have a great day!
<Step> 3/3
<Player> [action] go west. [words] None. [inventory]={}
<TextWorld> You are in the Main Street. To your north lies a library, to the west is the town hall, and to the south is a Blacksmith's shop.
<Game> location=main street
\end{lstlisting}

\subsection{Summarization Protocol}
The LLM inputs for generating game log summarizations include the prompt, in context examples, and finally the section of fresh game log for the LLM to summarize. In this section, we display several in-context examples. The prompt can be found in the released repository.

\label{appendix:summary}

\subsubsection{Selected In-context Examples}
\label{appendix:examples}
\begin{lstlisting}[numbers=none]
User:
Completed nodes so far: []
Sources: [Hatter location - NA, Bomb location - NA, Lab location - NA, Kit - NA]
<Step> 1/1
<Player> [action] take water bucket. [words] None. [inventory]={'water bucket'}
<TextWorld> You picked up the water bucket.
<Game> location=home
<Step> 2/2
<Player> [action] go west. [words] None. [inventory]={'water bucket'}
<TextWorld> You're in a quiet residential street. You see Mrs. Thompson walking her dog towards you. To the north, there's a park. To the west, there's a main street, and to the south, there's a restaurant.
<Game> location=residential street
<NPC> Mrs. Thompson: Hello!
<Step> 3/3
<Player> [action] None. [words] I'm trying to stop a bomb from exploding. [inventory]={'water bucket'}
<TextWorld> You're in a quiet residential street. You see Mrs. Thompson walking her dog towards you. To the north of you, there's a park. To the west, there's a main street, and to the south, there's a restaurant.
<Game> location=residential street
<NPC> Mrs. Thompson: A bomb? I don't know much. If you need help, ask the people around the village.
<Step> 4/4
<Player> [action] None. [words] Who should I ask about the explosion?. [inventory]={'water bucket'}
<TextWorld> You are in a residential street. You can see Mrs. Thompson walking her dog towards you. To the north of you, there's a park. To the west, there's main street, and to the south, there's a restaurant.
<Game> location=residential street
<NPC> Mrs. Thompson: Mad Hatter in the park might know something about it.

System:
[
{step: 1, total: 1, action: True, synopsis: Player adds water bucket to their inventory., success: True, node: [F1], helpful: True, completed nodes: []},
{step: 2, total: 2, action: True, synopsis: Player goes to residential street and meets Mrs. Thompson., success: True, node: [A1], helpful: True, completed nodes: [(A1,2)]},
{step: 3, total: 3, action: False, synopsis: Player tells Mrs. Thompson that they're trying to stop the bomb explosion., success: True, node: [A2], helpful: False, completed nodes: []},
{step: 4, total: 4, action: False, synopsis: Player asks Mrs. Thompson for who to talk to about the explosion and learns the Mad Hatter's location., success: True, node: [A2], helpful: True, completed nodes: [(A2,4)]}
]+sources: [Thompson, NA, NA, NA]+completed: [(A1,2),(A2,4)]

User:
Completed nodes so far: [(A1,2), (A2,15), (B1,16)]
Sources: [Hatter location - Thompson, Bomb location - NA, Lab location - NA, Kit - NA]
<Step> 17/17
<Player> [action] None. [words] Sure, I have a riddle for you. What stands on four legs in the morning, two legs in the afternoon, and three legs in the evening?. [inventory]={'redstone torch'}
<TextWorld> You're in the park. You see a man in a top hat standing near the statue. He appears to be muttering to himself and occasionally glancing around nervously. The exit is to the south.
<Game> location=park
<NPC> Mad Hatter: The answer is a human. What would you like to know from me?
<Step> 18/18
<Player> [action] None. [words] Mrs. Thompson said you know about the explosion that will happen later. [inventory]={'redstone torch'}
<TextWorld> You're in the park. You see a man in a top hat standing near the statue. He appears to be muttering to himself and occasionally glancing around nervously. The exit is to the south.
<Game> location=park
<NPC> Mad Hatter: I'll share some information, but you must solve this riddle first: "I'm a place where metal is shaped, and I have a hidden room where danger awaits. Where am I?" Solve this riddle, and you'll know the explosion location.
<Step> 19/19
<Player> [action] None. [words] Blacksmith shop?. [inventory]={'redstone torch'}
<TextWorld> You're in the park. You see a man in a top hat standing near the statue. He appears to be muttering to himself and occasionally glancing around nervously. The exit is to the south.
<Game> location=park;inventory={'redstone torch'}
<NPC> Mad Hatter: Bravo! The location of the explosion is the Blacksmith's shop storage room.

System:
[
{step: 17, total: 17, action: False, synopsis: Player tells the Mad Hatter a riddle to satisfy his request for information or a joke., success: True, node: [B2], helpful: True, completed nodes: []},
{step: 18, total: 18, action: False, synopsis: Player mentions the explosion to the Mad Hatter and receives a riddle from him about the bomb's location., success: True, node: [B2], helpful: True, completed nodes: [(B2,17)]},
{step: 19, total: 19, action: False, synopsis: Player solves riddle from the Mad Hatter about the bomb's location and the Mad Hatter confirms their answer., success: True, node: [B3, B4], helpful: True, completed nodes: [(B3,19), (B4,19)]}
]+sources: [Thompson, Hatter, NA, NA]+completed: [(A1,2), (A2,15), (B1,16), (B2,17), (B3,19), (B4,19)]
\end{lstlisting}

\subsubsection{Consistency Check of Summary}
\label{appendix:consistency}
We check the consistency of the log summaries along two axes. First, we ensure temporal consistency within each log. Second, we check that the summaries accurately reflect the gameplay. Below, we provide details for how each consistency check is performed.
\begin{enumerate}
\item The automated check for temporal consistency within each log is implemented as a Python script. It ensures that the summary of each log accurately reflects the sequence of events, preventing instances where scene "B" is reported as completed before scene "A," which should be completed first.
\item The authors conduct a consistency check between the summary and the original contents of the log to ensure accuracy. This involves verifying that game steps are correctly matched to scenes and that the single-sentence synopsis is accurate. This quality assurance step is performed by the authors to validate the effectiveness of our summarization protocol and ensure the robustness of our method. This is not part of the method, i.e. users need not conduct this check.
\end{enumerate}

\subsection{Additional Clustering Results}
\label{appendix:clusters}
Below are the clustering results for scenes identified as potential pain points in Fig \ref{fig:results}, not including those already discussed in Sec \ref{results}. 

\noindent Scene: Hatter asks riddle

\noindent Clusters:
\begin{enumerate}
    \item Player asks Mad Hatter about the bomb, but he doesn't provide any information.
    \item Player shares a riddle with the Mad Hatter and learns about the hidden bomb but not its location.
\end{enumerate}

\noindent Scenes: Maria reveals Lab in Library and Maria reveals Merlin in Lab

\noindent Clusters: 
\begin{enumerate}
    \item Player asks Maria about the hidden lab but doesn't get a clear answer.
    \item Player talks with Maria about the strange happenings in the village.
    \item Player asks Mrs. Thompson about the secret lab but she only confirms it's near the library.
    \item Player asks Maria for more information but receives the same information about Merlin.
    \item Player tells Maria they're trying to meet people and Maria warns them about Merlin.
\end{enumerate}

\noindent Scene: Meet Merlin

\begin{enumerate}
    \item Player asks Merlin to teach them magic, but Merlin requests proof that ingredient X is a lie.
    \item Player enters Merlin's lab but the game is reset before they can interact with him.
    \item Player goes to the secret lab and meets Merlin again.
    \item Player enters the secret lab and meets Merlin.
\end{enumerate}

\noindent Scene: Find storage room

\noindent Clusters: 
\begin{enumerate}
    \item Player tries to go to storage room but fails.
    \item Player tries to go south from the Blacksmith's shop but fails.
    \item Player looks for clues for the storage room but does not find anything.
    \item Player searches for secret rooms in Blacksmith’s shop but does not find anything.
    \item Player tries to go to the storage room but the game is reset before they can.
\end{enumerate}

\subsection{Ablation Study Prompts and In-context Examples}
\label{appendix:ablation}

\subsubsection{Naive Method}

\textbf{Prompt:}
\begin{lstlisting}[numbers=none]
You will be given transcripts from a game where the player is trapped in a time loop triggered by a bomb explosion and asked to help us find player pain points stemming from logical inconsistencies or too hard/easy pieces of the game design.

A <Game reset> occurs, starting a new round, each time the bomb explodes, which happens every 30 game steps.

The game town locations are: Home, Residential Street, Park, Restaurant, Main Street, Library, Town Hall, Blacksmith's Shop, Merlin's Lab, and the Storage Room.

The Player can encounter NPCs Chef Maria at the Restaurant, Mrs. Thompson in the Residential Street, Mad Hatter at the Park, James Moriarty in the Town Hall, and Merlin at Merlin's Lab.

The Player can put items in their inventory: water bucket from Home, redstone torch from Park, shears from Blacksmith's Shop, and journal from Library. There is a bomb disposal kit the player obtains by collecting these items and reading the journal or from Merlin.

To beat the game by disabling the bomb, players need to attain two main milestones: locate the bomb and acquire the bomb disposal kit.

To locate the bomb, they need to first convince Mrs. Thompson that they intend to stop the explosion, which will cause her to reveal that the Mad Hatter might have knowledge and that he is in the park.
The Mad Hatter will give the player a riddle which they must solve for him to reveal that the bomb is in the blacksmith's storage room. 
To acquire the bomb disposal kit, they have two options. The first option is to collect the water bucket, redstone torch, and shears, and read the recipe in the journal in the library. Alternatively, the player can find Merlin in the secret lab in the library and convince him to give them his disposal kit. In order to find Merlin, the player can either get Chef Maria to reveal that Merlin is in a secret lab in the library or learn of the secret lab's location from the map in the Town Hall.

The transcripts have three speakers:
<Player>: the words and actions of the player, along with the state of their inventory at the end of the current step
<TextWorld>: the feedback of the game world
<NPC>: a tag followed by the name of the NPC speaking and their speech

<Game> tells you the location of the player at the current game step, and additional feedback from the game world.
<Step> #/## means the player is on game step # in the current round and on game step ## cumulatively, including all previous rounds. You will say the current step number in this round for each step and the total number of steps so far.

Before the transcript section you will receive a brief summary of what has transpired so far in the game session.
You will list which parts of the game, if any, the player struggled with in the transcript.
You will also write a summary of the game session up until the end of the current section of transcript.
Include both of these components in a json-style dictionary.
\end{lstlisting}

\textbf{In-context Example:}
\begin{lstlisting}[numbers=none]
User:
Player created the bomb disposal kit from the ingredients and journal recipe. Player confronted James Moriarty with the information about Ingredient X and learned about the secret room in the library from the map.
<Step> 15/45
<Player> [action] None. [words] I am looking for the Mad Hatter. [inventory]={'shears', 'kit', 'redstone torch', 'water bucket'}
<TextWorld> lab
You are in a lab, filled with various chemical equipment and strange contraptions. In the corner, you see an old man hunched over a desk, scribbling furiously in a journal. As you approach, he looks up and greets you with a smile. You notice that he has a bomb disposal kit on his workbench. You can find the exit to the library by heading south.
<Game> location=lab
<NPC> Merlin: Ah, the Mad Hatter? He's quite an eccentric fellow, isn't he? I can't say I know his exact whereabouts at the moment, but I believe he enjoys spending time in the park. Perhaps you could try looking for him there?
<Step> 16/46
<Player> [action] go to the park. [words] None. [inventory]={'shears', 'kit', 'redstone torch', 'water bucket'}
<TextWorld> You are in the park. The sun is shining and birds are chirping. You notice a redstone torch placed on the bench. You can find the exit to the south.
<Game> location=park
<Step> 17/47
<Player> [action] None. [words] yell theres a bomb. [inventory]={'shears', 'kit', 'redstone torch', 'water bucket'}
<TextWorld> You are in the park. The sun is shining and birds are chirping. You notice a redstone torch placed on the bench. You can find the exit to the south.
<Game> location=park;feedback: You yell out, "There's a bomb!" Your voice echoes through the park, but there's no immediate response from anyone. It seems like you'll need to find more information about the bomb and its location before you can do anything about it.
<Step> 18/48
<Player> [action] None. [words] tell the eccentric crowd theres a bomb. [inventory]={'shears', 'kit', 'redstone torch', 'water bucket'}
<TextWorld> You are in the park. The sun is shining and birds are chirping. You notice a redstone torch placed on the bench. You can find the exit to the south.
<Game> location=park;feedback: You shout to the crowd of people in the park, "There's a bomb!" However, there is no eccentric crowd in the park at the moment. It appears that you need to find more information about the bomb and its location before you can take any action.
<Step> 19/49
<Player> [action] use the redstone torch. [words] None.  [inventory]={'shears', 'kit', 'redstone torch', 'water bucket'}
<TextWorld> You can't see any such thing.
<Game> location=park;feedback: You attempt to use the redstone torch, but there doesn't seem to be any purpose for it at the moment. You might need to find a specific situation or place where the redstone torch can be used effectively.
<Step> 20/50
<Player> [action] search in the bushes for the mad hatter. [words] None. [inventory]={'shears', 'kit', 'redstone torch', 'water bucket'}
<TextWorld> I only understood you as far as wanting to search inside.
<Game> location=park;feedback: You start searching the bushes in the park for any sign of the Mad Hatter. Unfortunately, you don't find him hiding in the bushes. You may need to look elsewhere or gather more information about his whereabouts.
<Step> 21/51
<Player> [action] go to the mad hatter. [words] None.  [inventory]={'shears', 'kit', 'redstone torch', 'water bucket'}
<TextWorld> You can't see any such thing.
<Game> location=park;feedback: You try to find the Mad Hatter in the park, but you can't see him at the moment. Perhaps you need more information about his whereabouts or a clue to his location.

System:
{"struggles": ["Finding the Mad Hatter"], "summary": "Player entered Merlin's lab and asked for the Mad Hatter's location, then tried to find the Mad Hatter in the park but failed."}
\end{lstlisting}

\subsubsection{No Logic Method}

\textbf{Prompt:}
\begin{lstlisting}[numbers=none]
You will be given transcripts from a game where the player is trapped in a time loop triggered by a bomb explosion and asked to help us find player pain points stemming from logical inconsistencies or too hard/easy pieces of the game design.

A <Game reset> occurs, starting a new round, each time the bomb explodes, which happens every 30 game steps.

The game town locations are: Home, Residential Street, Park, Restaurant, Main Street, Library, Town Hall, Blacksmith's Shop, Merlin's Lab, and the Storage Room.

The Player can encounter NPCs Chef Maria at the Restaurant, Mrs. Thompson in the Residential Street, Mad Hatter at the Park, James Moriarty in the Town Hall, and Merlin at Merlin's Lab.

The Player can put items in their inventory: water bucket from Home, redstone torch from Park, shears from Blacksmith's Shop, and journal from Library. There is a bomb disposal kit the player obtains by collecting these items and reading the journal or from Merlin.

To beat the game by disabling the bomb, players need to attain two main milestones: locate the bomb and acquire the bomb disposal kit.

To locate the bomb, they need to first convince Mrs. Thompson that they intend to stop the explosion, which will cause her to reveal that the Mad Hatter might have knowledge and that he is in the park.
The Mad Hatter will give the player a riddle which they must solve for him to reveal that the bomb is in the blacksmith's storage room. 
To acquire the bomb disposal kit, they have two options. The first option is to collect the water bucket, redstone torch, and shears, and read the recipe in the journal in the library. Alternatively, the player can find Merlin in the secret lab in the library and convince him to give them his disposal kit. In order to find Merlin, the player can either get Chef Maria to reveal that Merlin is in a secret lab in the library or learn of the secret lab's location from the map in the Town Hall.

James Moriarty had Merlin create the bomb in exchange for a made-up "Ingredient X" and had the Mad Hatter plant the bomb. Though he knows everything about the bomb, he has no power to disable the bomb or unlock locations/NPCs to the player.

The transcripts have three speakers:
<Player>: the words and actions of the player, along with the state of their inventory at the end of the current step
<TextWorld>: the feedback of the game world
<NPC>: a tag followed by the name of the NPC speaking and their speech

<Game> tells you the location of the player at the current game step, and additional feedback from the game world.
<Step> #/## means the player is on game step # in the current round and on game step ## cumulatively, including all previous rounds. You will say the current step number in this round for each step and the total number of steps so far.

Before the transcript section you will receive a list of which parts of the game the player has completed thus far.

For each step, you will summarize the player's underlying strategy or intention in that step, and if they successfully achieved it. You will tag the step with whether the player spoke or performed an action during the step.
You will assign each step to the part of the game that the player is trying to complete in that step.
For each step, you will say if the step was truly helpful in progressing towards the player's goal. 
For each step, you will write any newly completed parts of the game, along with the current total step number. 
These information should be catalogued in a json-style dictionary.

After each transcript segment, you will list all applicable completed sections of the game, along with the step number they were completed in, as context for the next segment. This should be a list of tuples.
\end{lstlisting}

\textbf{In-context Example:}
\begin{lstlisting}[numbers=none]
User:
Completed game parts so far: []
<Step> 1/1
<Player> [action] take water bucket. [words] None. [inventory]={'water bucket'}
<TextWorld> You picked up the water bucket.
<Game> location=home
<Step> 2/2
<Player> [action] go west. [words] None. [inventory]={'water bucket'}
<TextWorld> You're in a quiet residential street. You see Mrs. Thompson walking her dog towards you. To the north, there's a park. To the west, there's a main street, and to the south, there's a restaurant.
<Game> location=residential street
<NPC> Mrs. Thompson: Hello!
<Step> 3/3
<Player> [action] None. [words] I'm trying to stop a bomb from exploding. [inventory]={'water bucket'}
<TextWorld> You're in a quiet residential street. You see Mrs. Thompson walking her dog towards you. To the north of you, there's a park. To the west, there's a main street, and to the south, there's a restaurant.
<Game> location=residential street
<NPC> Mrs. Thompson: A bomb? I don't know much. If you need help, ask the people around the village.
<Step> 4/4
<Player> [action] None. [words] Who should I ask about the explosion?. [inventory]={'water bucket'}
<TextWorld> You are in a residential street. You can see Mrs. Thompson walking her dog towards you. To the north of you, there's a park. To the west, there's main street, and to the south, there's a restaurant.
<Game> location=residential street
<NPC> Mrs. Thompson: Mad Hatter in the park might know something about it.

System:
[
{"step": 1, "total": 1, "action": True, "summary": "Player adds water bucket to their inventory.", "success": True, "part": "Ingredient collection", "helpful": True, "completed": []},
{"step": 2, "total": 2, "action": True, "summary": "Player goes to residential street and meets Mrs. Thompson.", "success": True, "part": "Meeting Mrs. Thompson", "helpful": True, "completed": [("Meeting Mrs. Thompson",2)]},
{"step": 3, "total": 3, "action": False, "summary": "Player tells Mrs. Thompson that they're trying to stop the bomb explosion.", "success": True, "part": "Get the Mad Hatter's location", "helpful": False, "completed": []},
{"step": 4, "total": 4, "action": False, "summary": "Player asks Mrs. Thompson for who to talk to about the explosion and learns the Mad Hatter's location.", "success": True, "part": [Get the Mad Hatter's location], "helpful": True, "completed": [("Get the Mad Hatter's location",4)]}
]+completed: [("Meeting Mrs. Thompson", 2),("Get the Mad Hatter's location", 4)]
\end{lstlisting}

\subsection{Player Survey and Bug Comparison Details}
\label{appendix:ground-truth-comparison}

The player feedback form collected included only broad questions about player experience without asking players to specifically look for logic and game balance bugs. Additionally, players were not told the intended game design in detail, so the survey responses only reveal what players particularly noticed or disliked. To determine a ground truth from these surveys, we (human experts) manually parsed all survey responses and extracted responses indicating logic and/or game balance bugs, including both those responses which give a specific difficulty the player faced (e.g. finding the Mad Hatter) and those which give a general difficulty (e.g. experiencing some "non-functional" interactions).

We found 7 bugs reported by 11 of the 28 players. We computed the recall score, i.e. how many of these bugs were identified by our method, finding that 5 of these 7 bugs were detected by our method. The two bugs that were missed were (1) one player reported respawning at the Blacksmith's shop instead of in the Home and (2) two players reported that the kit failed to automatically assemble. However, upon applying the clustering part of our pipeline to the corresponding scenes, these two occurrences are caught: (1) is detected under the ``collect  bucket, shears, and torch’’ scene when the player tries and fails to pick up the water bucket -- this fails because they are in the Blacksmith's shop rather than the Home -- and (2) is detected in the ``get kit with journal’’ scene. Because these two scenes had overall high completion rates, we do not flag them as potential pain points in Stage 2 of our method, since the likelihood of these scenes being pain points to many players is low.

We also computed precision, though we note that precision is less informative of a metric since players were not explicitly asked to identify bugs, so not all players reported bugs. Of the 7 potential pain point scenes identified by our method, 5 were identified from the survey. The remaining two scenes concerned NPC Maria, whose design garnered general complaints in the survey. However, the survey responses do not cite specific interactions, so it is uncertain whether these complaints originated from the two pain point scenes our method identifies.

\subsection{Extended Ablation Results}
\label{appendix:ablation-results}

\subsubsection{Naive Method}
As described in Section~\ref{results:ablation}, the naive method (i.e. Method (1)), lacking game logic and summarization, structures creates difficulty in aggregating game parts across players and tracing back from game parts to player experience details. Here, we provide further examples of these difficulties. 

An example game log section summarization under Method (1) is as follows:
\begin{lstlisting}[numbers=none]
{'struggles': ['understanding the bomb disposal kit completion', 'interacting with merlin about ingredient x'], 'summary': 'the player struggled with understanding how to complete the bomb disposal kit and interacting with merlin regarding the secret ingredient x. they were told by merlin that the kit was incomplete without ingredient x, which led to confusion until merlin clarified that it was a metaphor for the combination of items the player already has or needs to find.'}
\end{lstlisting}

Below we provide examples of semantically equivalent game parts identified by Method (1) and the corresponding scene as defined by our game logic segmentation.

Scene: Meet Mrs. T
\begin{itemize}
    \item locating NPCs (Mrs. T)
    \item locating NPCs on Main Street
    \item interacting with Mrs. T
\end{itemize}

Scene: Meet Hatter
\begin{itemize}
    \item progressing towards finding the Mad Hatter
    \item locating the Mad Hatter after being told he is in the park
    \item locating the Mad Hatter despite asking around
    \item progressing towards finding the Mad Hatter despite being in the correct location (park)
    \item repeatedly trying to find the Mad Hatter without success
    \item player struggled to find the Mad Hatter, mistakenly asking Mrs. T if she was the Mad Hatter and not getting useful information on his whereabouts.
\end{itemize}

Below, we provide examples of overly vague or generic game parts.
\begin{itemize}
    \item locating NPCs
    \item potentially misunderstanding NPCs' hints
    \item progressing in the game
    \item proper use of obtained information
    \item understanding NPC responses
    \item lack of progress towards the main objective
\end{itemize}

Below, we provide examples of identified game parts that are highly specific to an individual player's gameplay and not a part of the intended game flow.
\begin{itemize}
    \item unable to order and pick up a spaghetti carbonara from Chef Maria
    \item giving the map to the Mad Hatter was not a correct action
    \item interacting with the wardrobe and its contents
    \item throwing the water bucket at Mrs. T was not a useful action
    \item understanding Merlin's reaction to the mention of Moriarty and Ingredient X without previously obtaining information about the secret room or the map in the town hall
    \item understanding the relationship between Merlin, Maria, and Ingredient X
\end{itemize}

In Table \ref{tab:naive-summaries}, we provide examples of several section summaries. These summaries often lack the detail for extracting specific obstacles people faced regarding a particular game section or are littered with additional information about the gameplay so far. Thus, even if the aggregation of game parts were resolved, mapping game parts back to specific player experiences proves challenging. Consequently, this method is unable to accurately provide reasons for any identified bugs.

\begin{table*}[!htbp]
\centering
\begin{tabular}{|p{3in}|p{3in}|}
\hline
\textbf{Game part identified} & \textbf{Summary of the specific player experience} \\
\hline
finding useful items in the restaurant, understanding the relevance of items in the restaurant & Player has progressed from the house to the residential street, interacted briefly with Mrs. T, and is now in the restaurant trying to find useful items. However, the player struggles to identify relevant items and has not made any significant progress towards stopping the bomb.\\
\hline
remembering game resets cause inventory loss & The player moved to the town hall, examined a map, and discovered a secret room in the library. they interacted with Moriarty, who inquired about their intentions. However, a game reset occurred, and the player lost all inventory items, starting a new round at their home. the player then decided to go to the blacksmith's shop.\\
\hline
understanding the game feedback prompting them to read the journal & Player obtained the journal in the library and was prompted to read it for more information. They also spoke to Merlin, who asked for help in finding components for the bomb disposal kit.\\\hline
\end{tabular}
\caption{Example game parts identified by Method (1) as parts players struggled with (left), along with a summary of the game steps each game part arose from (right). Since these summaries cover a section of the game log, it is challenging to automatically pinpoint the exact cause of struggle for each flagged game part. This renders extracting an actionable, concise list of bugs nearly impossible without manual parsing of the summaries.}
\label{tab:naive-summaries}
\end{table*}

In addition to these two issues, since this method seeks only to identify player pain points, the resulting summaries lack the information to determine game balance bugs where a section of the game is significantly easier than the designer intended.

\subsubsection{No Logic Method}

Like the naive method, the method lacking our game logic structure but using our summarization structure (Method (2)) also identifies many game parts which are unstandardized, too player specific, or too broad, making aggregation across players to find common obstacles difficult. However, we note that this method identifies fewer unique game part strings and, upon inspection, the game parts identified are easier to aggregate due to less varied phrasing and are more succinct than Method (1)'s.

An example gameplay step summarization under Method (2) is as follows:
\begin{lstlisting}[numbers=none]
{'step': 11, 'total': 11, 'action': True, 'summary': "Player goes to the Blacksmith's shop to search for bomb or ingredients.", 'success': True, 'part': "Navigating to Blacksmith's Shop", 'helpful': True, 'completed': [("Navigating to Blacksmith's Shop", 11)]}
\end{lstlisting}

Below, we provide several examples of identified game parts by Method (2).
\begin{itemize}
    \item Exploration
    \item Observation
    \item Investigating the bomb plot
    \item Interaction with Mrs. T's dog
    \item Acquiring bomb disposal kit
    \item Acquiring bomb disposal kit from Merlin
\end{itemize}

\begin{table*}[!htbp]
\centering
\begin{tabular}{|p{3in}|p{3in}|}
\hline
\textbf{Game part identified} & \textbf{Summary of the specific player experience} \\
\hline
Gathering information & Player ponders about useful items in the restaurant but doesn't take action. Maria suggests gathering more information.\\
\hline
Discovering Bomb Details & Player confronts Moriarty about his evil plans, but doesn't get any substantial information.\\\hline
Ingredient collection & Player attempts to take a bucket but fails due to the game reset.\\
\hline
Ingredient collection&Player takes the journal to learn about the bomb disposal kit.
\\\hline
Bomb disposal kit& Player attempts to store the bomb disposal recipe in their inventory, which is not a valid action.\\\hline
\end{tabular}
\caption{Example game parts identified by Method (2) as parts players struggled with (left), along with a summary of the game step each game part arose from (right).}
\label{tab:no-logic-summaries}
\end{table*}

In Table \ref{tab:no-logic-summaries}, we provide examples of several gameplay step summaries. These summaries provide greater detail for extracting specific obstacles players faced and more relevant to their corresponding game parts than those generated with Method (1). Thus, if the aggregation of game parts were resolved, mapping game parts back to specific player experiences to identify causes of bugs would be feasible.

Finally, since the summarization structure of this method maps gameplay steps back to attempted game parts, regardless of player struggle, it would allow for identification of ``too easy'' game balance bugs if the game part aggregation problem were solved.

\end{document}